\documentclass[smallabstract,smallcaptions]{dccpaper}

\usepackage{epsfig}
\usepackage{amsmath}
\usepackage{amssymb}
\usepackage{color}
\usepackage{siunitx}
\usepackage{makecell}
\usepackage{url}

\DeclareMathOperator*{\argmin}{arg\,min}
\usepackage{amsmath,amsfonts,bm}
\usepackage{caption,subcaption}
\usepackage[linesnumbered,ruled]{algorithm2e}
\usepackage[colorlinks=true]{hyperref}
\newlength{\figurewidth}
\newlength{\smallfigurewidth}
\begin{document}

\title
{\large
\textbf{Dataflow-based Joint Quantization of Weights and Activations for Deep Neural Networks}
}

\author{Xue Geng$^{\dag}$, Jie Fu$^{\ddag}$, Bin Zhao$^{\sharp}$, Jie Lin$^{\dag}$,
Mohamed M. Sabry Aly$^{\S}$, \\
Christopher Pal$^{\ddag}$, Vijay Chandrasekhar$^{\dag}$ \\
\begin{tabular}{cc}
$^{\dag}$I$^{2}$R, A$^{\star}$STAR & $^{\ddag}$MILA, IVADO\\
\footnotesize{\url{{geng_xue,lin-j,vijay}@i2r.a-star.edu.sg}}&\footnotesize{\url{{jie.fu,christopher.pal}@polymtl.ca}} \\
$^{\sharp}$IME, A$^{\star}$STAR & $^{\S}$School of CSE, NTU \\
\footnotesize{\url{zhaobin@ime.a-star.edu.sg}} & \footnotesize{\url{msabry@ntu.edu.sg}} \\
\end{tabular}
}
\maketitle
\thispagestyle{empty}
\begin{abstract}
This paper addresses a challenging problem -- how to reduce energy consumption without incurring performance drop when deploying deep neural networks (DNNs) at the inference stage. In order to alleviate the computation and storage burdens, we propose a novel dataflow-based joint quantization approach with the hypothesis that a fewer number of quantization operations would incur less information loss and thus improve the final performance. It first introduces a quantization scheme with efficient bit-shifting and rounding operations to represent network parameters and activations in low precision. Then it re-structures the network architectures to form unified modules for optimization on the quantized model. Extensive experiments on ImageNet and KITTI validate the effectiveness of our model, demonstrating that state-of-the-art results for various tasks can be achieved by this quantized model. Besides, we designed and 
synthesized an RTL model to measure the hardware costs among various quantization methods. For each quantization operation\footnote{Here, we choose to measure the energy consumption for various quantization methods on individual operations for simplicity. The overall energy consumption is in proportion to this measurement.}, it reduces area cost by ${\sim}15\times$ and energy consumption by ${\sim}9\times$, compared to a strong baseline. 
\end{abstract}

\Section{Introduction}
DNNs have been increasingly deployed at the edge due to its remarkable performance. In particular, energy becomes an import factor when processing DNNs at the edge in embedded devices with limited battery capacity (\textit{e.g.}, smartphones, smart sensors, UAVs, and wearables)~\cite{sze2017efficient}. Hence, there is an increasing demand for reducing energy consumption and increasing throughput without incurring a significant drop in accuracy.
Extensive research has focused on reducing the precision of operations and operands to address the challenge (\textit{e.g.}, DoReFa-Net~\cite{zhou2016dorefa}, BinaryNet~\cite{courbariaux2016binarized}, XNOR-Net~\cite{rastegari2016xnor} and ternary quantization~\cite{zhu2016trained}).

Although these quantization methods remarkably reduce the model complexity, they are limited in two key ways. Firstly, a noticeable performance drop exists. For example, DoReFa-Net has a 2.9\% performance loss for AlexNet on ImageNet with 8-bit weights and activations. Secondly, additional hardware costs are introduced due to extra inefficient operations such as codebooks in \cite{Han2015a} and scaling factors in DoReFa-Net and IOA~\cite{Jacob2017}. 


To address the above issues, we introduce an energy-efficiency quantization scheme which represents both network parameters and activations in low precision. It only contains efficient bit-shifting and rounding-to-nearest behaviors. Then, based on this scheme, we re-structure the network to form unified modules, which reduce the number of quantization operations. Here we hypothesize that fewer number of quantization operations would incur less information loss and thus boost the final performance accuracy. Finally, a joint reconstruction error loss function is set up on these unified modules to do optimization on the quantized model.

We evaluate our proposed method on ImageNet~\cite{deng2009imagenet} with ResNet~\cite{he2016deep} and KITTI~\cite{geiger2012we} with Faster R-CNN~\cite{ren2015faster}. Furthermore, we designed and synthesized a hardware unit to measure the hardware costs among various quantization operations. We show that our proposed approach performs comparably or even better in terms of accuracy and computational costs over existing state-of-the-art approaches on various tasks. In summary, our key contributions are as follows:
\begin{enumerate}
\item A purely quantized network is introduced to represent network parameters and activations into low precision with energy-efficient operations. As compared to floating-point representation, the 8-bit quantized model leads to less computation and memory accesses by ${\sim}4\times$ without significant performance drop.
\item A novel joint quantization approach is introduced by taking the network dataflow into consideration to define unified modules and then set up joint reconstruction error objectives to learn the optimal parameters for the quantization model. Besides, our approach does not have the time-consuming fine-tuning stage.
\item Extensive experiments on various benchmark datasets with state-of-the-art deep models demonstrate that our proposed method reduces computational costs while still achieving decent accuracy. 
\end{enumerate}

\section{Proposed Method}
In this section, we first describe our quantization scheme in DNNs and then provide a detailed induction on the integer-only inference. In the end, a joint quantization approach of network parameters and activations is presented.
\subsection{Quantization Scheme}
A common practice in reducing the precision of DNNs is to introduce a quantization function~\cite{zhou2016dorefa, Jacob2017}. A basic requirement of our quantization scheme is that it is hardware-friendly, \textit{i.e.}, without incurring intensive hardware operations and has equal conversion between integer and floating-point representation. The quantization scheme is defined as: Given a floating-point value $r$, we use a quantization function, $Q(\cdot)$, to approximate it:
\begin{align}
\small
\begin{split}
    r^q = Q(r; N_r, n_{bits}) =  \underbrace{min(2^{n_{bits} - 1} - 1, max(-2^{n_{bits} - 1}, round(r \times 2^{N_r})))}_{r^I} \times 2^{-N_r}
\end{split}
\end{align}
where $r^q$ is the quantized floating value, $r^I$ is the integer value and $N_r$ is the fractional bit which is the only parameter to set. $n_{bits}$ is the bit-width including $1$ sign bit. For example, when $n_{bits} = 8$, $r^I$ is a 8-bit integer within the range of $[-128, 127]$. When $N_r$ is negative, only the data before the decimal point is selected. For instance, if $N_r$ is -3 with 8-bit bit-width, we select the 3 to 10 digits before the decimal point as the low-precision data. Our quantization scheme only contains bit-shifting, which is different from previous works~\cite{zhou2016dorefa, Jacob2017} with multiplication-based scaling factors or \cite{ZhangYangYeECCV2018} weight loading from a codebook. 

In DNNs, a separate fractional bit is selected for weights, bias, and activations of each layer. For instance, a convolution layer would have a set of quantization parameters -- $N_w$, $N_b$, $N_x$, $N_o$ for weights, bias, input activation and output activation, respectively. $N_x$ comes from the output activation of the previous layer. 

\subsection{Integer-arithmetic-only Operations}
The above quantization scheme is in floating-point arithmetic and thus supposed to be deployed to general hardware platforms such as GPUs and CPUs. In this section, we describe an integer-only arithmetic inference in DNNs by adopting the proposed scheme. The inference provides a step-by-step solution on how custom hardware units compute on DNNs using fixed-point representations. 

Following the notations in \cite{sze2017efficient}, we define the 2-D convolution operation to be:
\begin{equation}
\small
\label{eq_float}
    \mathbf{O}_{l,m,n} = \mathbf{B}_{l} + \sum_{k=0}^{C-1} \sum_{i=0}^{H-1} \sum_{j=0}^{W-1} \mathbf{X}_{k,Sm + i, Sn+j}   \mathbf{W}_{l,k,i,j}
\end{equation}
where $\mathbf{O}$, $\mathbf{X}$, $\mathbf{W}$, and $\mathbf{B}$ are the matrices of the output feature maps (ofmaps), input feature maps (ifmaps), filters, and biases, respectively. $S$ is a provided stride while $C$, $H$ and $W$ are ifmaps channels, height and width, respectively. 

After quantizing weights, biases and activations, convolution operation becomes:
\begin{align}
\small
\label{eq_Fixed}
\begin{split}
    & CONV (\mathbf{X, W, B}; N_x, N_w, N_b) \\
    &= \mathbf{B}_{l}^I \times  2^{-N_b}  + \sum_{k=0}^{C-1} \sum_{i=0}^{H-1} \sum_{j=0}^{W-1} \mathbf{X}_{k,Sm + i, Sn+j}^I \times 2^{-N_x}  \times \mathbf{W}_{l,k,i,j}^I  \times 2^{-N_w} \\
    & = 2^{-N_w - N_x} (\mathbf{B}_l^I \times  2^{(N_x + N_w) - N_b} + \sum_{k=0}^{C-1} \sum_{i=0}^{H-1} \sum_{j=0}^{W-1} \mathbf{X}_{k,Sm + i, Sn+j}^I  \mathbf{W}_{l,k,i,j}^I) \\
    & = 2^{-N_w - N_x}  \times \mathbf{O\_int32}_{l,m,n} \\
\end{split}
\end{align}
where $\mathbf{X}^I$, $\mathbf{W}^I$ and $\mathbf{B}^I$ are the integer matrices of the ofmaps, ifmaps, filters, and bias, respectively. $N_x$, $N_w$ and $N_b$ are the fractional bit of ifmaps, filters, and biases, respectively. We carefully align biases with the convolution output by sacrificing smaller values achieving less information loss. The intermediate result of convolution is 32-bit integer to handle the overflow of accumulation.

Then, we need to quantize the output of the convolution layer as:
\begin{align}
\small
\begin{split}
\label{eq_q_output}
\mathbf{O}^q_{l,m,n} & = Q(CONV(\mathbf{X, W, B}; N_x, N_w, N_b), N_o, n_{bits})  \\
    &= \mathbf{O}_{l,m,n}^I \times 2^{-N_o}
\end{split}
\end{align}
where $\mathbf{O}^I$ is the integer matrices of the ofmaps. In custom hardware units, two sets of data are stored: one is the integer matrices of ifmaps $\mathbf{X}^I$, ofmaps $\mathbf{O}^I$, filters $\mathbf{W}^I$, and bias $\mathbf{B}^I$; and the other is the bit-shifting values for data alignment in biases and activations such as $N_x + N_W - N_b$ in Equation~\ref{eq_Fixed} but not the fractional bits. 

\subsubsection{Dataflow-based Joint Quantization}
\label{sec:hardware_friendly}
\begin{figure}[t]
\centering
\includegraphics[width=0.8\textwidth]{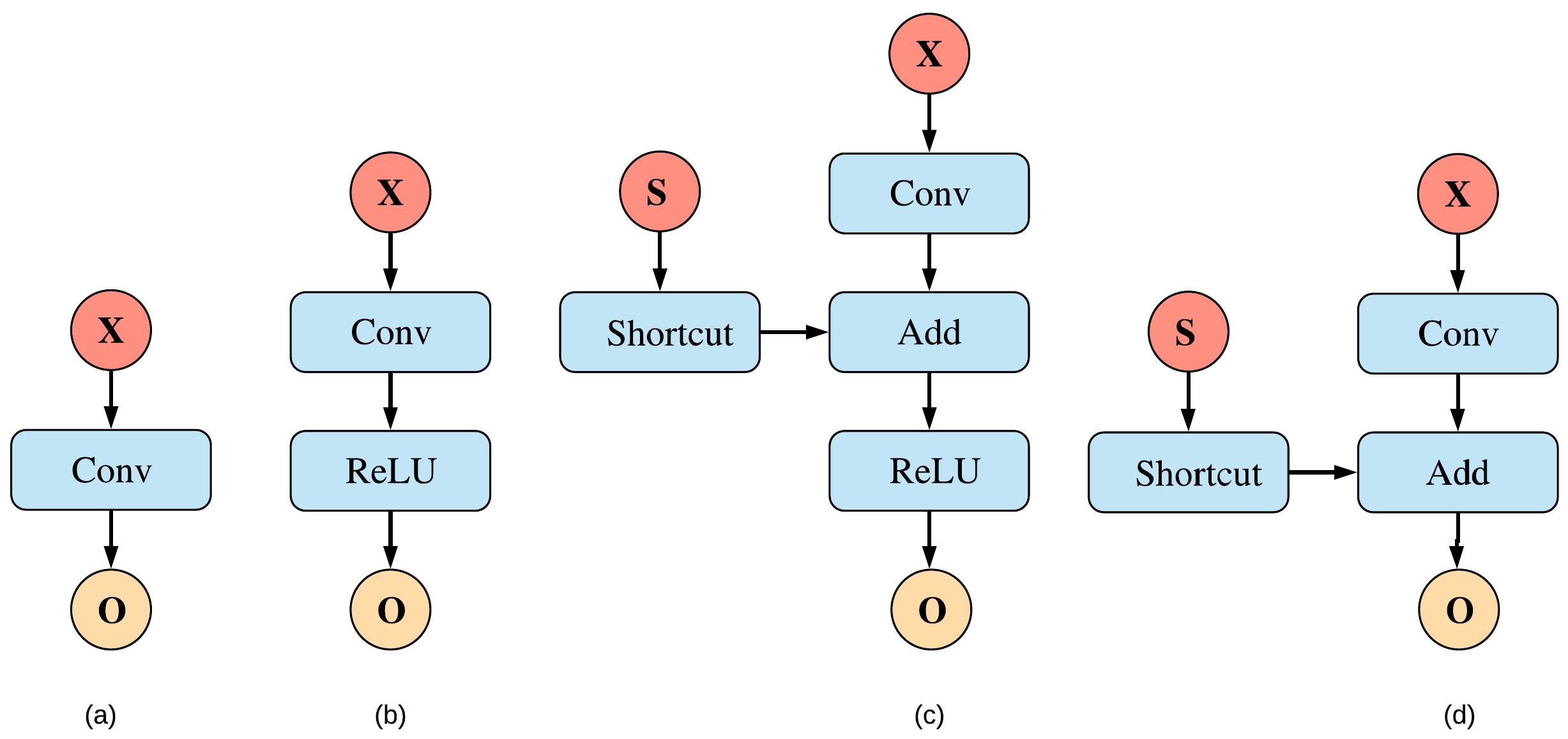}
\caption{In general, four common cases are considered here: a) convolution layer; b) a convolution layer followed by a ReLU layer; c) a residual connection with a ReLU layer; and d) a residual connection without a ReLU layer. }
\label{fig:quantization}
\end{figure}

The impact of activations on storage capacity depends on the network architecture and dataflow~\cite{sze2017efficient}. Hence, the dataflow of network architecture becomes an important factor to quantize the activations. Here we hypothesize that if more quantization operations are applied, it is likely to incur more measurement noise which might lead to information loss. To reduce the number of quantization operations, we propose to combine several basic layers into a unified module based on network architectures. 
We only consider ReLU as activation functions because it is the de facto choice for DNNs due to its effectiveness. 
For instance, the quantization is conducted after a ReLU layer in Figure~\ref{fig:quantization}(b) because that 1) the negative part can be skipped from computing the quantization error as only non-negative values exist after ReLU layer; 2) the cost of memory accesses is reduced dramatically without writing the convolution output back to memory. This is different from existing methods such as~\cite{zhou2016dorefa}, which quantizes activations instantly after convolution. Furthermore, the batch normalization layer is merged into the weights and biases of the next convolution layer at inference stage. 

Figure~\ref{fig:quantization} shows 4 cases to conduct joint quantization. In particular, Equation~\ref{eq_Fixed} is the simplest case in Figure~\ref{fig:quantization}(a). Figure~\ref{fig:quantization}(b) talks about the scenario of a convolution layer followed by ReLU, the activations are quantized after the ReLU layer instead of convolution layer. Thus the outputs of the ReLU layer is in the range $[0,255]$ if the bit-width is 8-bit. Figure~\ref{fig:quantization} (c) and (d) provide two situations of quantization on the residual connection, respectively. As Figure~\ref{fig:quantization}(c) has a residual connection from previous block, it is necessary to consider the alignment of shortcut outputs and convolution outputs. Furthermore, if the shortcut is a convolution layer, more complex alignment is done on two convolution layers. As compared to Figure~\ref{fig:quantization}(c), Figure~\ref{fig:quantization}(d) does not have a ReLU layer after addition. 

\subsubsection{Optimization}
Following \cite{ZhouMCF18}, we assume that the quantization error in each layer is positive relevant to the final performance accuracy. Hence it is reasonable to minimize each layer's reconstruction error to boost the performance. Our target is to minimize the reconstruction error between the outputs and the quantized outputs as follows:
\begin{equation}
\small
\label{eq_obj}
\hat{N}_w, \hat{N}_b, \hat{N}_o = \argmin_{N_w, N_b, N_o}||\mathbf{O} - Q(CONV(\mathbf{X, W, B}; N_x, N_w, N_b), N_o, n_{bits})||_2
\end{equation}
Where $\hat{N}_w, \hat{N}_b, \hat{N}_o$ are optimal fractional bits of weights, biases and output activations separately. ${N}_x$ is from the optimal bit of activations of the previous layer. 

\begin{algorithm}[!t]\footnotesize{
	\LinesNumbered
	\textbf{Initialization}:
	$\hat{N}_w \leftarrow 0$,
	$\hat{N}_o \leftarrow 0$,
	$\hat{N}_b \leftarrow 0$,
	$\hat{error} \leftarrow +\infty $,
	$n\_bits \leftarrow 8$,
	$\tau \leftarrow 4$.
	\\
	
	\textbf{Input}:
	$\mathbf{W}$,
	$\mathbf{B}$,
	$\mathbf{O}$.
	\\

	$N^{max}_{w} \leftarrow ceiling \left(\log_2\left(\max{(abs(\mathbf{W}))} + 1\right)\right) + 1, N^{min}_{w} \leftarrow N^{max}_w - \tau $ \\
	$N^{max}_{b} \leftarrow ceiling \left(\log_2\left(\max{(abs(\mathbf{B}))} + 1\right)\right) + 1,
	N^{min}_{b} \leftarrow N^{max}_b - \tau $ \\
	$N^{max}_{o} \leftarrow ceiling \left(\log_2\left(\max{(abs(\mathbf{O}))} + 1\right)\right) + 1,
	N^{min}_{o} \leftarrow N^{max}_o - \tau $ \\
	\ForEach{$i \in \left[N^{min}_{w}, N^{max}_w \right]$}{
	    $N_w = (n\_bits - 1) - i$, $\mathbf{W}^q \leftarrow Q \left(\mathbf{W}, N_w, n\_bits \right)$ \\
	    \ForEach{$j \in \left[N^{min}_b, N^{max}_b \right]$}{
	    $N_b = (n\_bits - 1) - j$, $\mathbf{B}^q \leftarrow Q (\mathbf{B}, N_b, n\_bits)$ \\
    	    \ForEach{$k \in [N^{min}_o, N^{max}_o]$}{
    	    $N_o = (n\_bits - 1) - k $, $\mathbf{O}^q \leftarrow Q (CONV(\mathbf{W}^q, \mathbf{B}^q, \mathbf{X}^q), N_o, n\_bits)$ \\
    	     $ error \leftarrow || \mathbf{O} -  \mathbf{O}^q||_2 $\\
    	    \If{$\hat{error} > error$}{
    	        $\hat{N}_w \leftarrow N_w$,  $\hat{N}_b \leftarrow N_b$, $\hat{N}_o \leftarrow N_o$, $\hat{error} \leftarrow error $
    	        }
    	    }
	    }
	}
	\KwRet{$\hat{N}_w, \hat{N}_b, \hat{N}_o$}
	\caption{$\textrm{Quantization on weights, biases and activations of convolution layer}$}
	\label{alg:quantization}
}
\end{algorithm}

Our proposed method is based on the pre-trained model without fine-tuning and the optimization is to learn the optimal fractional bits. To speed up the search, we first narrow down the search space and then apply a simple grid search approach. This is motivated by the observation~\cite{han2015learning} that larger weights play a more important role than smaller ones. Hence, we hypothesize that the optimal fractional bit should be located in the upper bits.  
Algorithm~\ref{alg:quantization} provides a detailed solution to the optimization. First, the largest fractional bit to represent the maximum value of parameter $\mathbf{W}$ is:
\begin{equation}
\small
 N^{max}_w = ceiling \left(\log_2\left(\max{(abs(\mathbf{W}))} + 1\right)\right) + 1
\end{equation}
The search space would be $[N^{max}_w, N^{max}_w - \tau]$ where $\tau$ is a hyper-parameter to be set empirically. Biases and activations have a similar search strategy. Now, we are ready to traverse all the solutions in a narrowed-down search space to get the optimal fractional bits. Obviously, the time complexity of Algorithm~\ref{alg:quantization} is $O(\tau^3 \Gamma)$ where $\Gamma$ comes from the convolution operation. In our experiments, we empirically set $\tau$ as 4. 

\section{Experiments}
We conduct experiments on ImageNet~\cite{deng2009imagenet}, KITTI~\cite{geiger2012we} datasets with two widely used deep models -- ResNet~\cite{he2016deep} and Faster R-CNN~\cite{ren2015faster} to investigate the performance of our proposed method. Besides, we conduct hardware measurements on various approaches to verify the computational efficiency of our proposed method. 
\subsection{Implementation Settings}
Our optimization is conducted on a single image as the number of weights, bias and activations for each layer are enough to bring little biases on the optimization. Meanwhile, the baselines are implemented according to their default settings. The CPU platform is AMD 1950X 16-Core CPU with 64GB RAM while the GPU platform is NVIDIA Tesla V100. 

To evaluate the hardware cost among various methods, we have created an RTL model for each method and conducted synthesis using UMC 40nm library, the area and power are then estimated at 500MHz clock Synthesizers. 

\subsection{Evaluation on ImageNet}
For training, all images are resized preserving aspect ratio so that the smallest side is 256. Then the center $224 \times 224$ patch is cropped and each of the RGB channels subtracts the means. While for inference, we apply a single-crop testing for standard evaluation. We apply our method to ResNet with different layer depths on ImageNet. The pre-trained floating-point model is from the TensorFlow's official repository \footnote{\label{footnote_1}\url{https://github.com/tensorflow/models/tree/master/research/slim}}. Method IOA~\cite{Jacob2017} is fine-tuned on the pre-trained floating-point model. 
\begin{table}
\centering
\caption{ResNet on ImageNet: Floating-point (FP) versus 8-bit quantized network for various network depths.}
\small
  \begin{tabular}{ c | c | c | c | c }
  \hline
  Methods & FP & TensorRT\cite{migacz20178} & IOA\cite{Jacob2017} & Ours \\ \hline
  ResNet-50 & 75.2\% & 73.1\% & 73.8\% & 73.6\%\\ \hline
  ResNet-101 & 76.4\% & 74.4\% & 74.6\% & 74.6\% \\ \hline
  ResNet-152 & 76.8\% & 74.7\% & 75.1\% & 75.0\% \\ \hline
  Quantization type & N/A  & scaling factor & scaling factor & bit-shifting \\ \hline
  \end{tabular}
 \label{tab:resnet}
\end{table}
\begin{table}
\centering
\caption{Training time for joint quantization for various network depths of ResNet.}
\small
  \begin{tabular}{ c | c | c | c }
  \hline
  Methods & ResNet-50 & ResNet-101 & ResNet-152 \\ \hline
  Training time (mins) & 5.6 & 7.1 &  8.5 \\ \hline
  \end{tabular}
\label{tab:resnet_time}
\end{table}
\begin{table}
\centering
\caption{ResNet50 on ImageNet: Accuracy under various approaches with different bit-widths.}
\small
  \begin{tabular}{ c | c | c | c | c | c }
  \hline
  Methods & CLIP-Q\cite{tung2018clip} & INQ\cite{zhou2017incremental} & ABC-net\cite{lin2017towards} & FGQ\cite{mellempudi2017ternary} & Ours \\ \hline
  Weight bits & 4 & 5 & 5 & 2 & 8 \\ \hline
  Activation bits & 32 & 32 & 5 & 8 &  8 \\ \hline
  Quantization type & codebook & N/A & scaling factor & scaling factor & bit-shifting \\ \hline
  Accuracy & 73.7\% & 74.8\% & 70.1\% & 70.8\% & 73.6\% \\ \hline
  \end{tabular}
\label{tab:resnet_baseline}
\end{table}

Table~\ref{tab:resnet} demonstrates the performance of ResNet-50, ResNet-101 and ResNet-152 on ImageNet. It can be seen that 1) our method is robust with various network depths with only about $1.8\%$ drop; 2) as compared to other baselines, our proposed approach achieves competitive performance. Although baseline IOA has similar results, it contains scaling factors and 32-bit biases. Besides, it has extra addition operations on the ``zero-point'' values. While our method saves more hardware cost with bit-shifting and 8-bit biases; and 3) the quantized model with optimal fractional bits is probably a great starting point to continue fine-tuning. It may further improve the performance and speed up the training as the search space of parameters becomes smaller.

Table~\ref{tab:resnet_time} lists the training time of our proposed method on ResNet with various network depths. We observe that in comparison with several-days fine-tuning on a pre-trained model, our approach has a dramatically fast speed within several minutes. Table~\ref{tab:resnet_baseline} reports the ImageNet accuracy for ResNet-50 on various approaches with various bit-widths. As expected, the integer-only work performs better than ABC-net, which uses 5-bit for weights and activations. INQ and CLIP-Q perform better than ours as they represent activations in 32-bit.  

\subsection{Evaluation on KITTI}
For object detection, we evaluate our method on the autonomous driving benchmark dataset -- KITTI, using Faster R-CNN (F-RCNN) with ResNet-152 as the backbone network. Since the ground-truth for test set is not publicly available, we randomly sample 80\% of training images for training F-RCNN and the remaining for validation on the quantized model. All images are resized into HD resolution ($1280 \times 720$). 
\begin{table}
\centering
\caption{Object detection performance on KITTI dataset for various data precision.}
\small
  \begin{tabular}{ c | c | c | c | c }
  \hline
   & FP & 8-bit & 7-bit & 6-bit \\ \hline
  Car & 94.35\% & 93.98\% & 92.32\% & 75.30\% \\ \hline
  Pedestrian & 80.78\% & 81.43\% & 79.06\% & 53.85\% \\ \hline
  Cyclist & 87.63\% & 86.87\% & 85.34\% & 48.05\% \\ \hline
  \end{tabular}
  \label{tab:kitti}
\end{table}

Table~\ref{tab:kitti} lists the object detection performance of our proposed method in terms of various bit-widths on KITTI dataset.  We observe that our method in 8-bit achieves almost and even better performance as compared to full precision models. Our method in 7-bit precision also has a competitive performance. However, 6-bit representation has a dramatic performance drop. 

\subsection{Hardware Experiments}
We measure the energy and area costs in Table~\ref{tab:hw} on various operations. To do comparison fairly, all implementations are constrained to 32-bit input and 8-bit output. In particular, scaling factor operation is implemented by a 32-bit multiplier and the output is then clipped to the rightmost 8-bit integer value; k-means has a 4-bit codebook with each entry a 8-bit value. The selected entry is multiplied with input data and then clipped to the rightmost 8-bit; while for bit-shifting, the input data is shifted right by the range in [1,10] and then clipped the rightmost 8-bit.

\begin{table}
\centering
\caption{Hardware cost of various methods with 32-bit inputs and 8-bit outputs.}
\small
  \begin{tabular}{ c | c | c | c}
  \hline
   Operation types& scaling factor & codebook & bit-shifting \\ \hline
   Power (mW) & 30.6 & 228.8 & 15.5 \\ \hline
   Area (\SI{}{\micro\metre^2}) & 502.7 & 1787.6 & 198.2 \\ \hline
  \end{tabular}
  \label{tab:hw}
\end{table}


It shows that bit-shifting saves the power and area the most. 
The scaling factor operation consumes ${\sim}2\times$ more energy and area of bit-shifting operation. The codebook consumes the power and area most as the codebook contains intensive encoding-decoding operations. Following~\cite{howard2017mobilenets}, if in floating-point precision, the computational cost of the quantization for activations is about $\frac{1}{\text{filter size}}$ of the standard convolution layer, only occupying about $1-2\%$ of the whole network computation. However, in fixed-point precision, the computational cost of quantization cannot be ignored. For example, for the scaling factor experiment, the convolution layer is implemented by 8-bit multipliers while the quantization is 32-bit multipliers, the computational cost of quantization would be increased by ${\sim}16\times$ and should not be ignored. 


\subsection{Discussion}
\label{sec:hardware}
\begin{figure}
\centering
\small
\begin{subfigure}{0.39\textwidth}
\includegraphics[width=\linewidth]{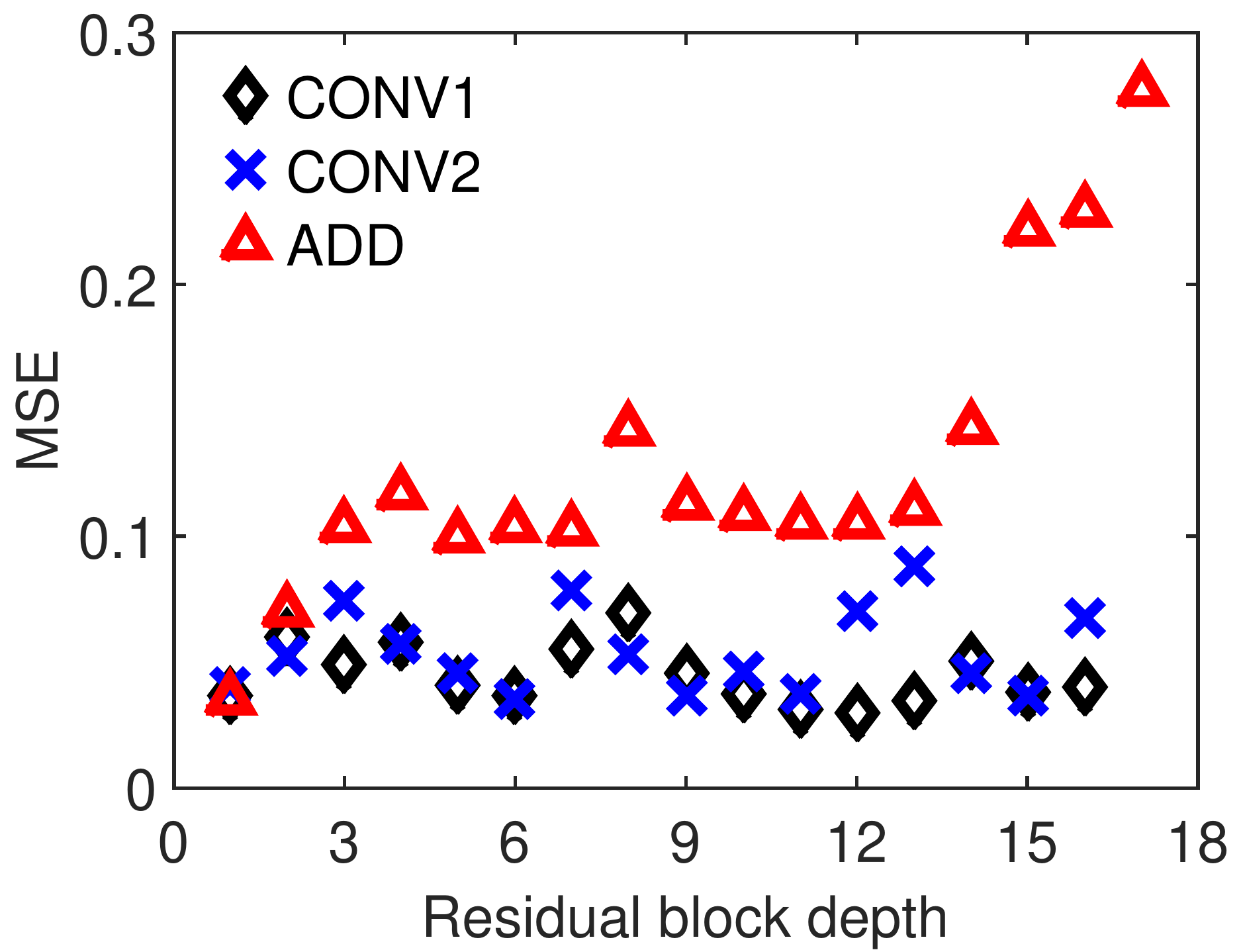}
\caption{MSE with residual block depth.} \label{fig:quantization_error}
\end{subfigure}
\hspace{8mm} 
\begin{subfigure}{0.40\textwidth}
\includegraphics[width=0.9\linewidth]{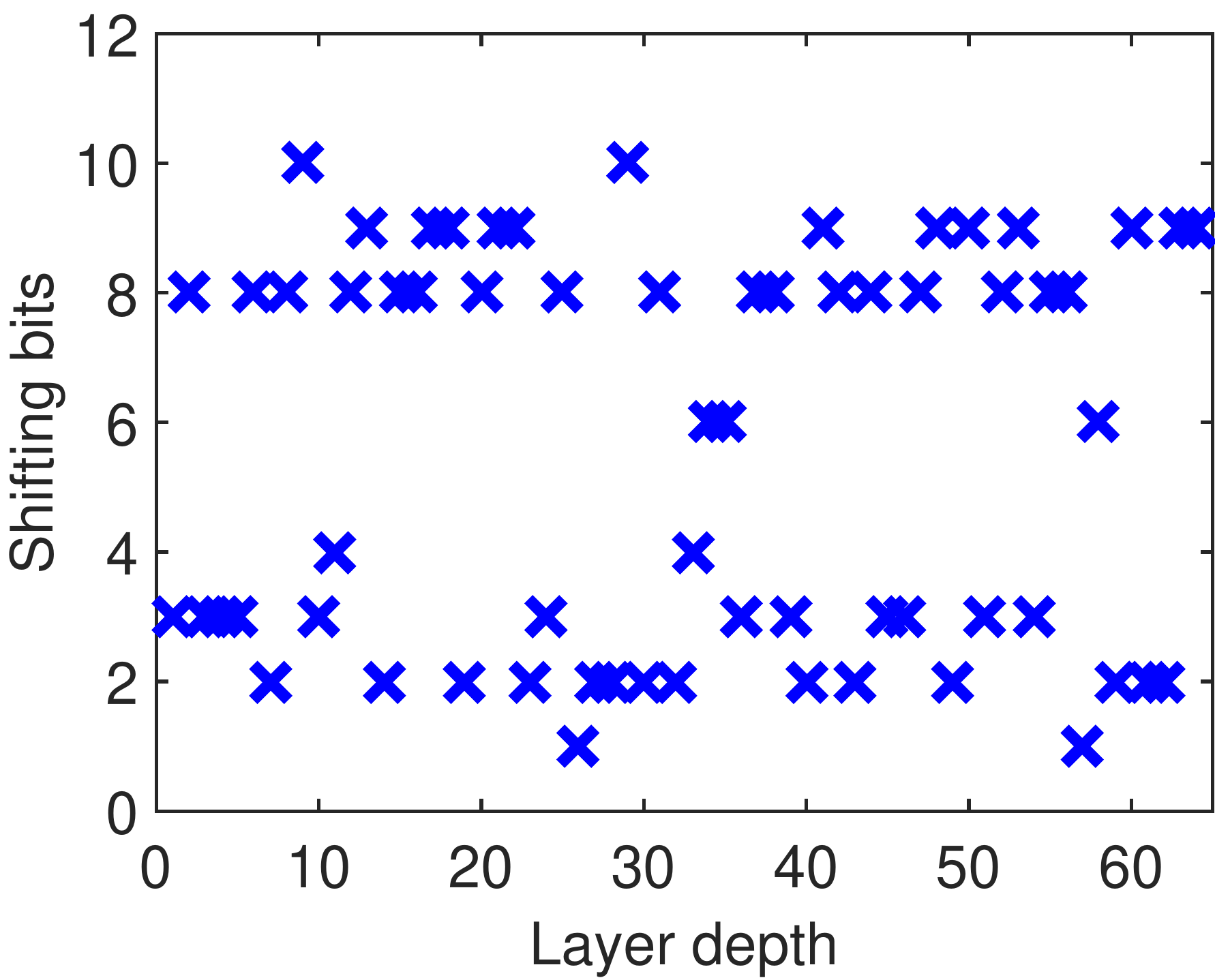}
\caption{Shifting bits with layer depth.} \label{fig:shift_bits}
\end{subfigure}
\caption{Statistics on ResNet-50.}
\label{fig:stats}
\end{figure}

Figure~\ref{fig:stats} shows the statistics on mean squared error (MSE) between quantized activations and floating-point activations with residual block depth and the shifting bits with layer depth for ResNet-50. From Figure~\ref{fig:quantization_error}, we observe that 1) the MSE of residual addition is larger than the first two convolution layers as it integrates last convolution layer and shortcut connection; and 2) when the layer goes deeper, the MSE error of the first two convolutions in a residual block is stable while the addition units increase. From Figure~\ref{fig:shift_bits}, it can be seen that 1) the bit-shifting operation operates in a range [1,10]; and 2) the bit-shifting values often revolves around 3 and 8, respectively.

\section{Related Work}
Recent work on compressing DNNs while accelerating inference can be classified as: a) adopting low-precision of both operands and operations; and b) reducing the number of operations by designing compact architectures or pruning.

\textbf{Lower precision} Some works have been focused on reducing the precision of weights for efficient on-chip memory storage~\cite{zhou2017incremental, zhu2016trained}. For instance, \cite{zhou2017incremental} proposes an incremental network quantization (INQ) to convert pre-trained full-precision models into low-precision ones  by quantizing weights into power of two or zero values. To further save the computing cost and memory storage, recent works also consider the quantization of activation~\cite{zhou2016dorefa,courbariaux2016binarized,rastegari2016xnor,Xu2018}. For instance, binarized neural networks (BNNs)~\cite{courbariaux2016binarized} uses binary weights and activations and thus reduces the MAC to an XNOR operation. Different from these methods, a novel dataflow-based joint quantization approach is proposed to quantize the weights and activations.

\textbf{Fewer operations} There is a significant amount of research on designing efficient network architectures~\cite{iandola2016squeezenet, howard2017mobilenets, Chollet_2017_CVPR}. SqueezeNet~\cite{iandola2016squeezenet} uses $1\times1$ instead of $3\times3$ filters, which dramatically increases the depth of DNNs thus reducing the model complexity.
Xception~\cite{Chollet_2017_CVPR} and MobileNet~\cite{howard2017mobilenets} rely on depthwise separable convolutions.
Yet another way to reduce the number of operations is through network pruning~\cite{Han2015a, Yang_2017_CVPR, NIPS2015_5784}. As an example, \cite{NIPS2015_5784} prunes the network based on the magnitude of weights. As the model size of a DNN does not directly reflect the hardware energy consumption, \cite{Yang_2017_CVPR} proposes an energy-aware method to prune the weights. 

\Section{Conclusions}
In this work, we explored a dataflow-based joint quantization mechanism to lower the precision of network parameters and activations in DNNs and performed extensive experiments to demonstrate the effectiveness of our proposed methods in terms of performance accuracy and hardware costs. The quantized model is a purely low-precision model with weights, biases and activations in low precision. Besides, the batch normalization is merged into the convolution layer. 

\Section{References}
\bibliographystyle{IEEEbib}
\bibliography{refs}
\end{document}